\pdfoutput=1

\documentclass[11pt]{article}

\usepackage[final]{acl}

\usepackage{times}
\usepackage{latexsym}

\usepackage[T1]{fontenc}

\usepackage[utf8]{inputenc}
\usepackage{microtype}
\usepackage{inconsolata}
\usepackage{graphicx}
\usepackage{hyperref}

\usepackage{xspace}

\usepackage{enumitem}
\usepackage{amsmath}
\usepackage{booktabs}
\usepackage{multirow}

\setlist{leftmargin=*,nosep}

\title{TathyaNyaya and FactLegalLlama: Advancing Factual Judgment Prediction and Explanation in the Indian Legal Context}

\author{Shubham Kumar Nigam$^{1,4*\dagger}$ \quad Balaramamahanthi Deepak Patnaik$^{1*}$ \\
\textbf{Shivam Mishra}$^{1}$ \quad 
\textbf{Noel Shallum}$^{2}$ \quad \textbf{Kripabandhu Ghosh}$^{3}$ \quad \textbf{Arnab Bhattacharya}$^{1}$\\
$^{1}$Indian Institute of Technology Kanpur, India \quad
$^{2}$Symbiosis Law School Pune, India \\
$^{3}$IISER Kolkata, India \quad
$^{4}$University of Birmingham, Dubai, United Arab Emirates\\
\texttt{\{shubhamkumarnigam, bdeepakpatnaik2002,}
\texttt{shivam1602m, noelshallum\}@gmail.com} \\ 
\quad \texttt{kripaghosh@iiserkol.ac.in} \quad \texttt{arnabb@cse.iitk.ac.in}
}

\date{}

\begin{document}
\maketitle

{
\renewcommand{\thefootnote}{$*$}
\footnotetext{These authors contributed equally to this work}
\renewcommand{\thefootnote}{$\dagger$}
\footnotetext{Corresponding author}
\renewcommand{\thefootnote}{\arabic{footnote}}
}

\begin{abstract}
In the legal domain, Fact-based Judgment Prediction and Explanation (FJPE) aims to predict judicial outcomes and generate grounded explanations using only factual information, mirroring early-phase legal reasoning. Motivated by the overwhelming case backlog in the Indian judiciary, we introduce \textbf{\texttt{TathyaNyaya}}, the first large-scale, expert-annotated dataset for FJPE in the Indian context. Covering judgments from the Supreme Court and multiple High Courts, the dataset comprises four complementary components, \texttt{NyayaFacts}, \texttt{NyayaScrape}, \texttt{NyayaSimplify}, and \texttt{NyayaFilter}, that facilitate diverse factual modeling strategies.
Alongside, we present \textbf{\texttt{FactLegalLlama}}, an instruction-tuned LLaMa-3-8B model fine-tuned to generate faithful, fact-grounded explanations. While \texttt{FactLegalLlama} trails transformer baselines in raw prediction accuracy, it excels in generating interpretable explanations, as validated by both automatic metrics and legal expert evaluation. Our findings show that fact-only inputs and preprocessing techniques like text simplification and fact filtering can improve both interpretability and predictive performance. 
Together, \texttt{TathyaNyaya} and \texttt{FactLegalLlama} establish a robust foundation for realistic, transparent, and trustworthy AI applications in the Indian legal system.
\end{abstract}

\section{Introduction}
The integration of AI technologies into the legal domain holds immense potential for improving the efficiency, accessibility, and transparency of judicial processes, particularly in countries like India, where case backlogs severely burden the courts. As of recent estimates, over 50 million cases are pending across various courts in India~\citep{njdc-district}, resulting in delays that can stretch into decades. In this context, early-phase legal decision support, i.e., prediction based solely on factual information available at the beginning of a case, has emerged as a highly relevant research goal.
Among the emerging solutions, Fact-based Judgment Prediction and Explanation (FJPE) offers a promising direction. FJPE aims to predict judicial outcomes and provide rationales using only the factual elements of a case, without relying on arguments, precedents, or judicial reasoning. This mirrors real-world scenarios where stakeholders, judges, lawyers, or litigants, must assess case strength based on initial facts to decide whether to proceed, allocate resources, or pursue alternative legal remedies. Furthermore, factual records are often the most reliably documented and readily available components in early legal proceedings, especially in resource-constrained environments.

While previous studies have attempted fact-centric modeling by summarizing multiple legal components or relying on automatically extracted facts~\cite{nigam-etal-2024-rethinking, 10.1145/3632754.3632765}, these approaches often lack reliable ground truth and blur the boundaries between pure factual inputs and broader legal discourse. Moreover, such works typically reference the full case context, statutes, or reasoning, placing them closer to the domain of CJPE, which includes post-filing evidence and legal argumentation. In contrast, FJPE distinctly isolates factual segments to simulate the setting of early-phase legal reasoning, where preliminary decisions may be formed even before formal hearings begin.
To advance this direction, we introduce \texttt{TathyaNyaya}, the first large-scale, expertly annotated dataset explicitly designed for FJPE in the Indian legal context. The term combines the Hindi words ``Tathya'' (fact) and ``Nyaya'' (justice), underscoring its foundation in factual legal analysis. Unlike prior datasets, \texttt{TathyaNyaya} does not rely on heuristics or summarization techniques; instead, it offers cleanly annotated factual inputs aligned with judicial outcomes and explanations, allowing for reproducible, interpretable, and practical early-stage prediction models.

\texttt{TathyaNyaya} comprises judgments from the Supreme Court of India (SCI) and various High Courts and is organized into four components: \texttt{NyayaFacts}, \texttt{NyayaScrape}, \texttt{NyayaSimplify}, and \texttt{NyayaFilter}. These components support a wide range of fact-centric tasks, from expert annotations and simplified factual paraphrasing to fact vs. non-fact segmentation.
Complementing the dataset, we introduce {FactLegalLlama}, an instruction-tuned version of LLaMa-3-8B, fine-tuned on {TathyaNyaya} to perform FJPE tasks. While transformer-based models are strong in predictive performance, {FactLegalLlama} demonstrates the ability to generate faithful and interpretable factual explanations, thus bridging predictive modeling with legal reasoning.

Our key contributions in this paper are:
\begin{itemize}
    \item \emph{{TathyaNyaya} Dataset:} We introduce the first extensively annotated, purely fact-centric dataset for judgment prediction and explanation in the Indian legal domain, structured into four components tailored for factual segmentation, simplification, and retrieval.
    
    \item \emph{Early-Phase Legal Reasoning:} We focus on realistic and societally impactful early-phase decision-making settings where predictions are made using only the facts, reflecting constraints and needs of India's overburdened judiciary.

    \item \emph{{FactLegalLlama} for Explanation:} We propose \texttt{FactLegalLlama}, an instruction-tuned LLaMa-3-8B model designed to generate faithful and fact-grounded explanations for judicial outcomes. It excels in producing coherent and semantically aligned rationales.
\end{itemize}

Our dataset, code, and model are available through a GitHub repository\footnote{\href{https://github.com/ShubhamKumarNigam/TathyaNyaya-and-FactLegalLlama}{TathyaNyaya-and-FactLegalLlama GitHub Repo}}.

\section{Related Work}
\label{sec:related-work}

Judgment Prediction has evolved into one of the most studied problems in legal NLP, driven by the need for computational assistance in judicial decision-making. Foundational studies such as \citet{aletras2016predicting}, \citet{chalkidis2019neural}, and \citet{feng2021recommending} first demonstrated the feasibility of predicting legal outcomes from textual case records, leading to benchmark datasets such as CAIL2018 \cite{xiao2018cail2018} and ECHR-CASES \cite{chalkidis2019neural}. These works inspired architectures that integrate predictive performance with interpretability, setting the foundation for subsequent developments.

In the Indian context, significant efforts have been made to address judgment prediction and related downstream tasks. The ILDC corpus \cite{malik-etal-2021-ildc} provided the first large-scale annotated dataset for CJPE, while later initiatives such as \texttt{NyayaAnumana} \cite{nigam2024nyayaanumana} and PredEx \cite{nigam2024legaljudgmentreimaginedpredex} expanded interpretability by combining factual and reasoning-based perspectives. Subsequent research explored complementary directions, including Legal Question Answering (AILQA) \cite{nigam2023legal}, rhetorical role segmentation for legal document structuring \cite{malik2021semantic, nigam2025legalseg}, and structured document generation with \texttt{VidhikDastaavej} \cite{nigam2025structured}. Further, \citet{vats-etal-2023-llms, nigam2022nigam} analyzed the efficacy of Large Language Models (LLMs) such as GPT and LLaMA for statute and judgment prediction, exposing both potential and bias in legal AI. 
Recent datasets like \texttt{NyayaRAG} \cite{nigam2025nyayarag}, and IBPS \cite{srivastava2025ibps} continue advancing realistic, explainable, and fact-grounded AI frameworks within the Indian judicial system.

Fact-based LJP has emerged as a more interpretable and realistic paradigm, focusing on early-phase factual understanding. Works such as \citet{nigam-etal-2024-rethinking} and \citet{10.1145/3632754.3632765} explicitly emphasized the challenge of reasoning solely from facts while maintaining legal coherence. However, those approaches relied on extractive or summarization-based proxies for factual representation. Our work directly addresses this gap by providing expert-annotated factual data in \texttt{NyayaFacts} and by fine-tuning \texttt{FactLegalLlama} to generate faithful, fact-grounded explanations. 
This focus aligns with the broader explainability and interpretability movement in legal NLP, where models are expected to provide transparent, human-understandable justifications \cite{marino2023automatic, santosh2024hiculr}.

Beyond India, multilingual and cross-jurisdictional research has broadened the global reach of LJP. The SwissJudgmentPrediction dataset \cite{niklaus2021swiss} and HLDC corpus \cite{kapoor-etal-2022-hldc} extended analysis to non-English and Hindi legal texts, respectively. Similarly, rhetorical and event-based modeling approaches \cite{feng-etal-2022-legal, santosh2025lecopcrlegalconceptguidedprior} enriched semantic structure understanding. 
Parallel initiatives like the SAIL Symposium \cite{ghosh2023report, ganguly2023legal} have consolidated research in Legal IR, information retrieval, and reasoning tasks.

\section{Task Description}
Our work centers on predicting and explaining legal judgments from the Supreme Court of India (SCI) and various High Court cases using a newly introduced annotated dataset, \texttt{TathyaNyaya}. This dataset is the largest of its kind for factual judgment prediction and explanation in the Indian legal domain. Unlike prior approaches relying on full case texts, \texttt{TathyaNyaya} emphasizes factual information alone, reflecting more realistic conditions for automated legal decision-making.

We divide \texttt{TathyaNyaya} documents into 2 sets:
\begin{itemize}
    \item \textbf{Single:} Either it contains a single petition or multiple petitions where all decisions are identical. 
    \item \textbf{Multi:} It contains multiple appeals with different outcomes. For simplicity, we convert all {partially accepted} cases into {accepted}, preserving the binary classification setup. Thus, both \texttt{single} and \texttt{multi} datasets support binary classification.
\end{itemize}

The task consists of two subtasks:

\textbf{Task A: Judgment Prediction:} This is a binary classification problem. Given the factual information of a legal case, the goal is to predict whether the judgment favors the appellant or not. A label of "1" denotes acceptance (including partially accepted cases), and "0" denotes complete rejection.

\textbf{Task B: Rationale Explanation:} This subtask involves generating a textual explanation for the predicted decision. The rationale should be grounded in the provided factual information and reflect the reasoning that supports the outcome.

Figure~\ref{fig:fjpe_task_framework} in the Appendix illustrates the overall FJPE pipeline, outlining the stages from fact input to prediction and explanation generation.

\section{Dataset}
\label{sec:dataset}

\begin{figure*}[t]
    \centering
    \includegraphics[width=0.7\textwidth, height=0.65\textwidth]{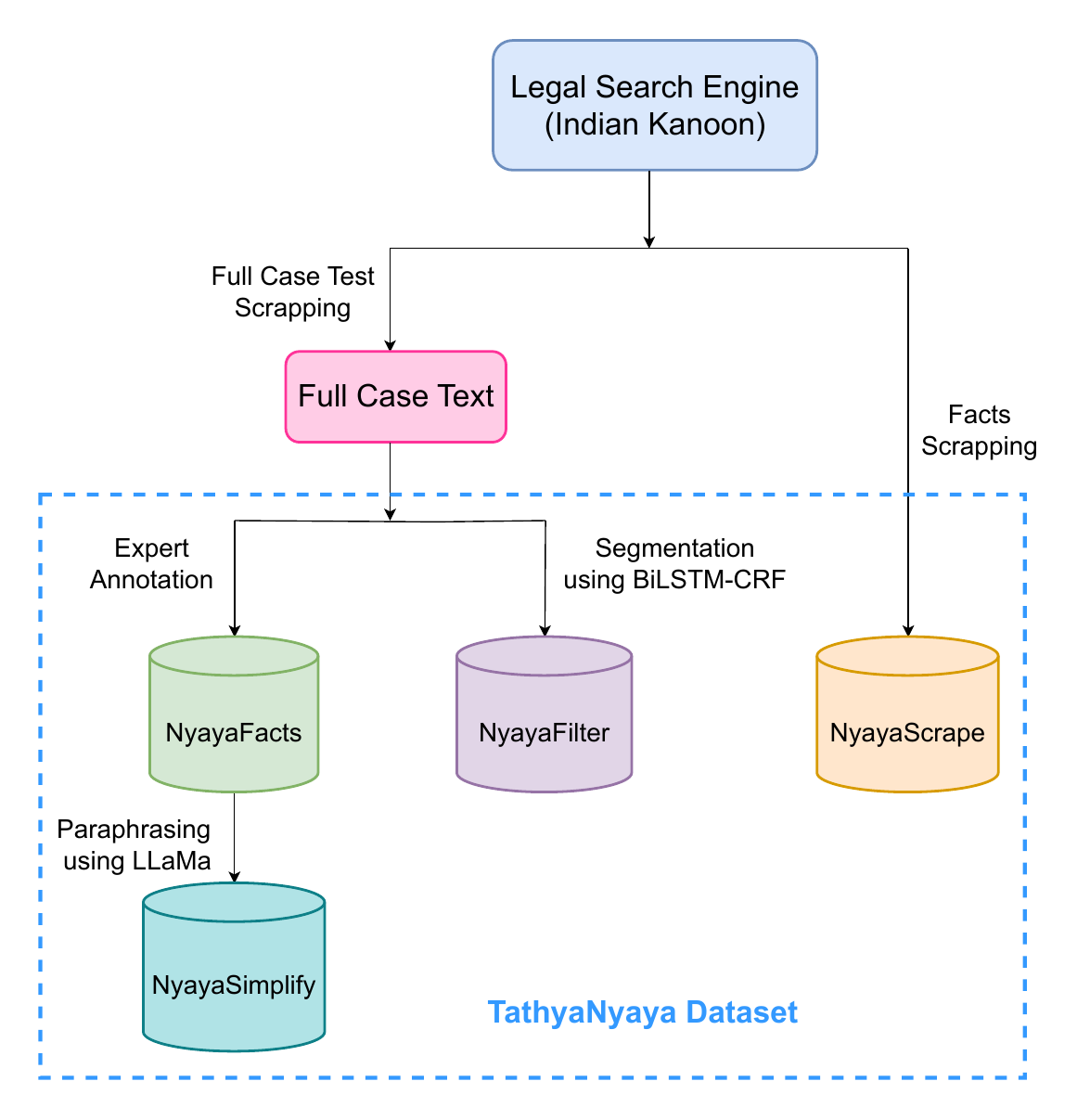}
    \caption{A high-level illustration of the \texttt{TathyaNyaya} dataset creation pipeline, showcasing the development process and interconnections of its four components.}
    \label{fig:tathya_pipeline}
\end{figure*}
In this research, we introduce \texttt{TathyaNyaya}, a comprehensive dataset explicitly designed for FJPE in the Indian legal domain. This dataset consists of four distinct components: (1) \texttt{NyayaFacts}: expert-annotated data that serves as the gold standard for prediction and explanation tasks, (2) \texttt{NyayaScrape}: automated fact-extracted data obtained through machine-driven processes, (3) \texttt{NyayaSimplify}: a user-friendly dataset created by paraphrasing complex legal language, and (4) \texttt{NyayaFilter}: a binary fact vs. non-fact classification dataset designed to streamline the retrieval of relevant factual information. Together, these components form the largest and most diverse factual dataset in the Indian judiciary, enabling the development and evaluation of advanced AI models for transparent and interpretable judgment prediction and explanation. By focusing exclusively on factual data, \texttt{TathyaNyaya} addresses a critical gap in the field, paving the way for more robust and realistic AI-driven solutions tailored to the Indian legal context.

Figure~\ref{fig:tathya_pipeline} illustrates the \texttt{TathyaNyaya} dataset creation pipeline. It provides a high-level overview of how each component which is derived, from expert-curated facts and machine-driven extraction, to fact segmentation and paraphrasing. This end-to-end pipeline ensures that the final dataset captures both breadth and depth in factual legal information.

\subsection{Dataset Compilation and Statistics}
The compilation process involved collecting approximately 16,000 judgments from the Supreme Court of India (SCI) and various High Courts through IndianKanoon\footnote{\url{https://indiankanoon.org/}}, a widely used legal search engine known for its comprehensive repository of Indian legal documents. These judgments were then categorized into the following components:

\subsubsection{\texttt{NyayaFacts}} \texttt{NyayaFacts} comprises a subset of different Court judgments carefully annotated by legal experts. These annotations highlight key factual segments that significantly influence judicial outcomes, serving as high-quality ground truth for both judgment prediction and rationale explanation. After refining and preprocessing, this subset serves as the gold standard for the FJPE task.

In particular, the validation and test data were derived from the \texttt{NyayaFacts Single} subset to maintain consistency during evaluation, while the training data include both single and multi-case judgments, offering a broad learning landscape. Table~\ref{tab:dataset-stats} provides comprehensive statistics. 
\begin{table}[t]
\centering
\resizebox{\linewidth}{!}{%
\begin{tabular}{lcccc}
\toprule
\textbf{Metric} & \textbf{Train (Multi)} & \textbf{Train (Single)} & \textbf{Validation} & \textbf{Test} \\
\midrule
\multicolumn{5}{c}{\textbf{NyayaFacts}} \\ 
\midrule
\# Documents & 13,629 & 8,216 & 1,197 & 2,389 \\
Avg \# Words & 855 & 853 & 828 & 865 \\ 
Acceptance (\%) & 55.20 & 47.66 & 47.45 & 47.72 \\ 
\midrule
\multicolumn{5}{c}{\textbf{NyayaScrape}} \\ 
\midrule
\# Documents & 8,993 & 3,828 & 548 & 1,095 \\ 
Avg \# Words & 405 & 404 & 412 & 405 \\
Acceptance (\%) & 65.77 & 61.44 & 59.85 & 60.55 \\ 
\bottomrule
\end{tabular}
}
\caption{Statistics for \texttt{NyayaFacts} and \texttt{NyayaScrape} datasets from the \texttt{TathyaNyaya} corpus.}
\label{tab:dataset-stats}
\end{table}

\subsubsection{\texttt{NyayaScrape}} \texttt{NyayaScrape} comprises judgments sourced from the Indiankanoon website, where cases are automatically segmented into various categories such as facts, issues, conclusions, and assessments of how the courts have treated certain elements (e.g., "Negatively Viewed by Court," "Relied by Party," "Accepted by Court"). Although these segments aim to provide structured insights, the labels are not entirely reliable. They are generated by automated tools rather than human legal experts, resulting in potential inconsistencies and may introduce noise. Moreover, not all judgments contain every type of label, further complicating the data’s uniformity.

Despite these limitations, \texttt{NyayaScrape} offers valuable machine-derived factual extractions that enable us to compare expert-driven annotations with automated processes. This comparison helps assess the reliability, quality, and shortcomings of model-based fact identification and segmentation. Document-level statistics and comparisons against \texttt{NyayaFacts} are provided in Table~\ref{tab:dataset-stats}.


\subsubsection{\texttt{NyayaSimplify}}
\texttt{NyayaSimplify} aims to enhance model performance and interpretability by transforming complex legal texts into simplified, paraphrased versions. Since most LLMs are pre-trained on general-purpose corpora and not on legal-specific jargon, they often struggle with the dense and domain-specific language found in court judgments. To address this, we paraphrased the \texttt{NyayaFacts} test data using the LLaMA-3-70B-Instruct model. This transformation preserves the factual and legal integrity of the original content while expressing it in more accessible, human-readable language.

The resulting dataset allows us to evaluate whether simplifying legal language helps general-purpose models better understand and reason about legal facts. While most dataset statistics remain consistent with \texttt{NyayaFacts}, the average word count is notably reduced, indicating a successful simplification. Our findings suggest that simplification improves both the accuracy and interpretability of models on FJPE tasks. Prompt template used for paraphrasing is included in Appendix Table~\ref{tab:nyayasimplify_prompt}.

\subsubsection{\texttt{NyayaFilter}} 
\texttt{NyayaFilter} addresses the challenges of manual annotation by employing a BiLSTM-CRF model to classify sentences as either factual (1) or non-factual (0). This binary classification replaces the traditional multi-label approach, simplifying the task while maintaining a focus on essential factual information. The model was trained on \texttt{NyayaFacts Single} data, with validation and testing on the corresponding splits. This approach achieved approximately 90\% accuracy in separating factual statements, as shown in Table~\ref{tab:filter-stats}. This dataset streamlines the retrieval process for FJPE tasks and enables scalable fact extraction.

\begin{table}[t]
\centering
\resizebox{\linewidth}{!}{%
\begin{tabular}{lrrr}
\toprule
\textbf{Metric} & \textbf{Train} & \textbf{Validation} & \textbf{Test} \\ 
\midrule
\multicolumn{4}{c}{\textbf{Facts}} \\ 
\midrule
\# Documents & 13,629 & 1,197 & 2,389 \\
\# Sentences & 3,62,658 & 30,561 & 56,240 \\
Avg \# Words & 29.00 & 29.00 & 34.00 \\
Avg \# Facts/Document (\%) & 23.6 & 23.03 & 22.7 \\
Overall Facts (\%) & 19.16 & 19.09 & 18.46 \\
\midrule
\multicolumn{4}{c}{\textbf{Non-Facts}} \\ 
\midrule
\# Documents & 13,629 & 1,197 & 2,389 \\
\# Sentences & 15,29,998 & 1,29,543 & 2,48,433 \\
Avg \# Words & 28.00 & 28.00 & 30.00 \\
Avg \# Non-Facts/Document (\%) & 76.4 & 76.97 & 77.3 \\
Overall Non-Facts (\%) & 80.84 & 80.91 & 81.54 \\ 
\bottomrule
\end{tabular}
}
\caption{Comparison of factual vs. non-factual statistics used during BiLSTM-CRF classifier training for the \texttt{NyayaFilter} dataset.}
\label{tab:filter-stats}
\end{table}

\subsection{Annotation Process \& Quality Assurance}
\subsubsection{Expert Participation}
The annotation process for \texttt{NyayaFacts} was carried out by a team of 10 legal experts, comprising advanced third- and fourth-year law students from premier Indian law colleges. These individuals were chosen based on their academic standing, legal reasoning skills, and familiarity with judicial processes, ensuring that the annotations reflected high-quality and domain-relevant insights.

\subsubsection{Timeline and Workload Distribution}
The annotation process was conducted over an extended period (April 1, 2022, to October 30, 2023), reflecting the complexity and precision required to analyze diverse legal texts. Each annotator was assigned approximately 30 judgment documents per week, a volume that balanced efficiency with attention to detail. This measured pace allowed the annotators to thoroughly examine the factual segments without compromising quality.

\subsubsection{Annotation Protocol}
The annotators were tasked with identifying and extracting specific judgment segments that contained factual information, without personal interpretation or summarization. This approach preserved the authenticity of the annotations, ensuring that they faithfully represented the judicial reasoning within each document.

\subsubsection{Quality Control Framework}
To maintain annotation consistency and reliability, a multi-layered quality control mechanism was implemented:
\begin{itemize}[leftmargin=*] 
    \item \textbf{Initial Review:} Each case was initially annotated by a single expert. This ensured efficiency while maintaining focus on factual segments. Subsequently, the annotations underwent multiple validation layers.
    
    \item \textbf{Senior Expert Validation:} Discrepancies or ambiguous annotations were escalated to a review panel comprising senior legal practitioners, who provided final judgments on contentious segments, enhancing the reliability of the final annotations.
    
    \item \textbf{Training and Alignment Meetings:} Regular training sessions and coordination meetings were conducted to align all annotators on annotation protocols, legal conventions, and factual identification criteria. These interactive forums helped minimize subjectivity, solidify common standards, and maintain uniform annotation quality throughout the project’s duration.

    \item \textbf{Framework Evaluation and Reproducibility Assessment:} To assess the stability and reproducibility of the factual annotation process, we conducted an independent evaluation involving two additional legal annotators. A randomly selected subset of 50 documents from the \texttt{NyayaFacts} dataset was re-annotated using the same annotation guidelines. The original annotations were treated as the reference (gold), and textual similarity metrics, BLEU, METEOR, ROUGE-1/2/L, and BERTScore were computed to compare the new annotations with the original.

\begin{table}[t]
\resizebox{\linewidth}{!}{%
    \centering
    \begin{tabular}{lcccccc}
        \toprule
        \textbf{Annotator Pair} & \textbf{BLEU} & \textbf{METEOR} & \textbf{R1} & \textbf{R2} & \textbf{RL} & \textbf{BERTScore} \\
        \midrule
        A2 vs A1 (Gold) & 0.78 & 0.62 & 0.85 & 0.68 & 0.80 & 0.91 \\
        A3 vs A1 (Gold) & 0.75 & 0.60 & 0.83 & 0.65 & 0.78 & 0.89 \\
        \bottomrule
    \end{tabular}}
    \caption{Agreement scores between original annotations (A1) and two additional annotators (A2 and A3) across 50 re-annotated legal cases.}
    \label{tab:annotation-agreement}
\end{table}

    These results demonstrate strong agreement between the independent annotators and the original gold annotations. In particular, high ROUGE-1 ($>$0.80) and BERTScore ($>$0.89) values indicate consistent extraction of factually relevant segments, both lexically and semantically. This empirical evidence supports the reproducibility and reliability of the annotation framework employed for \texttt{NyayaFacts}, validating its use as a high-quality benchmark for fact-based legal judgment prediction and explanation.
    
\end{itemize}

\section{Methodology}
\label{sec:methodology}
In this section, we present our overall methodology for extracting factual segments from legal judgments, training our custom model \texttt{FactLegalLlama} for FJPE, and finally addressing both the prediction-only and prediction-with-explanation tasks. We also detail the prompts we used and instruction-tuning strategies employed to refine our model’s outputs.

\subsection{Fact Extraction from Full Judgments}
\label{subsec:fact_extraction}
To prepare the dataset for FJPE, we first extracted the factual statements from full-text legal judgments. We adopted a streamlined binary classification approach by fine-tuning a BiLSTM-CRF model~\cite{ghosh2019identification}, a previous state-of-the-art (SoTA) model for semantic segmentation of legal documents. Instead of using the original multi-class rhetorical role framework, which distinguishes between roles such as issue, statute, precedent, and argument, we simplified the task by treating all non-factual segments as a single class labeled "non-facts".
This transformation into a binary classification problem enabled the model to focus solely on identifying factual segments critical to judgment prediction. Training was conducted using the \texttt{NyayaFacts} multi, which provided expert-annotated labels for factual and non-factual segments. By isolating the facts, we laid the groundwork for developing AI models capable of making decisions and generating explanations based solely on factual data. This preprocessing ensured that the subsequent models trained on the dataset remained focused on the most relevant and actionable information in legal cases.

\subsection{Training \texttt{FactLegalLlama}}
\label{subsec:factlegalllama_training}
The \texttt{FactLegalLlama} model, based on LLaMa-3-8B architecture, was fine-tuned specifically for the FJPE task using \texttt{NyayaFacts}. The training process involved instruction-tuning with a diverse set of 16 templates designed to guide the model in judgment prediction and explanation tasks. We utilized low-rank adaptation (LoRA) to optimize model training on limited computational resources. Training parameters, such as quantization to 4-bit precision and gradient accumulation, ensured efficient usage of resources while maintaining model performance.
To further enhance its capabilities, {FactLegalLlama} was fine-tuned with both prediction-only and with explanation tasks, enabling it to handle a wide range of factual judgment scenarios. 

\subsection{Fact-Based Judgment Prediction}
\label{subsec:judgment_prediction}
\subsubsection{Language Model-Based Approach}
For baseline comparisons, we utilized transformer-based models like InLegalBERT~\cite{paul-2022-pretraining}, and XLNet Large~\cite{yang2019xlnet} for binary classification. Due to the token length constraints of these models, we adopted a chunking strategy by dividing documents into 512-token segments with a 100-token overlap to preserve context. Chunk-level predictions were aggregated to generate final case-level predictions.

\subsubsection{Large Language Model-based Approach}
We utilized \texttt{FactLegalLlama}, our instruction-tuned LLaMa-3-8B model~\cite{dubey2024llama}, for judgment prediction-only instructions, where the model predicts judicial outcomes solely based on the factual inputs. The training data from \texttt{TathyaNyaya} was used to train the factual prediction context, emphasizing precision.

\subsection{Prediction with Explanation (FJPE)}
\label{subsec:judgment_prediction_explanation}
For the combined task of prediction and explanation, we employed \texttt{FactLegalLlama} with modified instruction prompts. Instructions guided the model to first predict the outcome and then generate a rationale grounded in the provided factual data.

\subsection{Prompts Used}
\label{subsec:prompts_templates}
Prompts for both prediction and explanation tasks were carefully designed the prompts. For prediction-only tasks, the prompts instructed the model to output a binary decision. For prediction-with-explanation tasks, the prompts included directives to explain the reasoning behind the prediction. Templates are detailed in Table~\ref{tab:factual_judgment_prediction_prompts} in the Appendix.

\subsection{Instruction Sets}
\label{subsec:instruction_tuning}
The fine-tuning process for \texttt{FactLegalLlama} involved using a diverse set of 16 instruction templates for judgment prediction and explanation. These templates ensured the model could generalize effectively across a wide range of cases and factual scenarios. The complete list of instruction sets used for tuning is in Table~\ref{Instruction-sets} in the Appendix.

\section{Evaluation Metrics}
\label{sec:performance_metrics}

To rigorously assess the performance of our models on judgment prediction and factual explanations in the \texttt{TathyaNyaya} test dataset, we employed a suite of evaluation metrics. For judgment prediction, we report Macro Precision, Recall, F1, and Accuracy. For evaluating the quality of explanations, both quantitative and qualitative methods were applied.

\begin{enumerate}
    \item \textbf{Lexical-Based Evaluation:} We used traditional lexical similarity metrics, including ROUGE-1/2/L \cite{lin-2004-rouge}, and BLEU \cite{papineni-etal-2002-bleu}. These metrics measure word overlap and sequence alignment between generated explanations and reference texts, providing a quantitative measure of the accuracy of lexical content.

    \item \textbf{Semantic Similarity Evaluation:} To assess the semantic alignment of the generated explanations, we applied BERTScore \cite{BERTScore}, which evaluates semantic similarity between the generated text and reference explanations. Additionally, BLANC \cite{blanc} was utilized to estimate the contextual relevance and coherence of the generated text in the absence of a gold-standard reference.

    \item \textbf{Expert Evaluation:} To validate the interpretability and legal soundness of the model-generated explanations, we conduct an expert evaluation involving legal professionals. They rate a representative subset of the generated outputs on a 1–10 Likert scale across three criteria: factual accuracy, legal relevance, and completeness of reasoning. A score of 1 denotes a poor or misleading explanation, while a 10 reflects high legal fidelity and argumentative soundness. This evaluation provides critical insights beyond automated metrics.

    \item \textbf{Inter-Annotator Agreement (IAA):} To further ensure the reliability and consistency of expert judgments, we computed multiple standard inter-rater agreement metrics. Specifically, Fleiss’ Kappa~\cite{fleiss1971measuring} was used to evaluate the overall agreement among multiple raters, Cohen’s Kappa~\cite{cohen1960coefficient} captured pairwise agreement while adjusting for chance, and the Intraclass Correlation Coefficient (ICC)~\cite{shrout1979intraclass} measured the reliability of continuous ratings across raters. In addition, Krippendorff’s Alpha~\cite{krippendorff2018content} provided a robust measure suitable for ordinal scales and missing data, while the Pearson Correlation Coefficient~\cite{benesty2009pearson} quantified the linear consistency of expert scores. The results demonstrated substantial agreement across these different measures, reinforcing the credibility of the evaluation and lending strong support to the robustness of our findings.
\end{enumerate}


\section{Results and Analysis}
\label{sec:results_analysis}
\begin{table}[t]
\centering
\resizebox{\linewidth}{!}{%
\begin{tabular}{lccccl}
\toprule
\textbf{Model}       & \textbf{\begin{tabular}[c]{@{}c@{}}Macro \\ Precision \end{tabular}} & \textbf{\begin{tabular}[c]{@{}c@{}}Macro \\ Recall \end{tabular}} &
  \textbf{\begin{tabular}[c]{@{}c@{}}Macro \\ F1 \end{tabular}} &
  \textbf{\begin{tabular}[c]{@{}c@{}}Accuracy \end{tabular}} & \textbf{\begin{tabular}[c]{@{}c@{}}Training \\ Data \end{tabular}}  \\ \midrule
\multicolumn{6}{c}{\textbf{Results on NyayaFacts Test Data}}                                                                                           \\ \midrule
InLegalBert          & 0.5934                   & 0.5936                & 0.5935            & 0.5932            &  \\ 
XLNet\_Large         & \textbf{0.6064}                   & \textbf{0.6040}                & \textbf{0.6052}            & \textbf{0.6061}            &                              \\ 
FactLegalLlama       & 0.5416                   & 0.5312                & 0.5036            & 0.5386            &     \multirow{-3}{*}{\begin{tabular}[c]{@{}l@{}}NyayaFacts\\ Single\end{tabular}}                         \\ \midrule
InLegalBert          & 0.6001                   & 0.5836                & 0.5917            & 0.5740            &   \\ 
XLNet\_Large         & \textbf{0.6145}                   & \textbf{0.5965}                & \textbf{0.6054}            & \textbf{0.5908}            &                              \\ 
FactLegalLlama       & 0.5390                        & 0.5368                     & 0.5318                 & 0.5401                 &      \multirow{-3}{*}{\begin{tabular}[c]{@{}l@{}}NyayaFacts\\ Multi\end{tabular}}             \\ \midrule
InLegalBert          & 0.5480                   & 0.5192                & 0.5332            & 0.5082            &   \\ 
XLNet\_Large         & \textbf{0.5807}                   & \textbf{0.5781}                & \textbf{0.5794}            & \textbf{0.5756}            &                              \\ 
FactLegalLlama       & 0.5139                        & 0.5122                     & 0.4922                 & 0.5042                 &   \multirow{-3}{*}{\begin{tabular}[c]{@{}l@{}}NyayaScrape\\ Single\end{tabular}}                            \\ \midrule
InLegalBert          & 0.5735                   & 0.5269                & 0.5492            & 0.5157            &    \\ 
XLNet\_Large         & \textbf{0.5935}                   & \textbf{0.5878}                & \textbf{0.5906}            & \textbf{0.5842}            &                              \\ 
FactLegalLlama       & 0.4951                        & 0.4966                     & 0.4516                 & 0.4884                 & \multirow{-3}{*}{\begin{tabular}[c]{@{}l@{}}NyayaScrape\\ Multi\end{tabular}}                             \\ \midrule 
\multicolumn{6}{c}{\textbf{Results on NyayaScrape Test Data}}                                                                                           \\ \midrule 
InLegalBert          & 0.6718                   & 0.5748                & 0.6195            & 0.6521            &  \\ 
XLNet\_Large         & \textbf{0.6754}                   & \textbf{0.6394}                & \textbf{0.6569}            & \textbf{0.6849}            &                              \\ 
FactLegalLlama       & 0.5574                        & 0.5372                     & 0.5191                 & 0.6045                 &   \multirow{-3}{*}{\begin{tabular}[c]{@{}l@{}}NyayaScrape\\ Single\end{tabular}}                            \\ \midrule
InLegalBert          & 0.7976                   & 0.7268                & 0.7606            & 0.7717            &   \\ 
XLNet\_Large         & \textbf{0.8098 }                  & \textbf{0.7781}                & \textbf{0.7936}            & \textbf{0.8055}            &                              \\ 
FactLegalLlama       & 0.5439                        & 0.5317                     & 0.5177                 & 0.5877                 & \multirow{-3}{*}{\begin{tabular}[c]{@{}l@{}}NyayaScrape\\ Multi\end{tabular}}                             \\ \midrule
InLegalBert          & \textbf{0.6237}                   & 0.5243                & 0.5697            & \textbf{0.6183}            &  \\ 
XLNet\_Large         & 0.5433                   & 0.5282                & 0.5357            & 0.5918            &                              \\ 
FactLegalLlama       & 0.5832                        & \textbf{0.5868}                     & \textbf{0.5792}                 & 0.5840                 &  \multirow{-3}{*}{\begin{tabular}[c]{@{}l@{}}NyayaFacts\\ Single\end{tabular}}                             \\ \midrule
InLegalBert          & \textbf{0.6784}                   & 0.5027                & 0.5775            & 0.6073            & \\ 
XLNet\_Large         & 0.6124                   & 0.5129                & 0.5583            & 0.6119            &                              \\ 
FactLegalLlama       & 0.6541                        & \textbf{0.6583}                     & \textbf{0.6552}                 & \textbf{0.6651}                 &  \multirow{-3}{*}{\begin{tabular}[c]{@{}l@{}}NyayaFacts\\ Multi\end{tabular}}                             \\ \bottomrule
\end{tabular}%
}
\caption{Performance metrics of models evaluated on NyayaFacts and NyayaScrape test data. Each block shows results obtained by training on either NyayaFacts or NyayaScrape data (single or multi variants), then testing on corresponding subsets.}
\label{tab:merged_results_test}

\end{table}

\begin{table}[t]
\centering
\resizebox{\linewidth}{!}{%
\begin{tabular}{lccccl}
\toprule
\textbf{Model} & \textbf{\begin{tabular}[c]{@{}c@{}}Macro \\ Precision\end{tabular}} & \textbf{\begin{tabular}[c]{@{}c@{}}Macro \\ Recall\end{tabular}} & \textbf{\begin{tabular}[c]{@{}c@{}}Macro \\ F1\end{tabular}} & \textbf{\begin{tabular}[c]{@{}c@{}}Accuracy\end{tabular}} & \textbf{\begin{tabular}[c]{@{}c@{}}Training \\ Data\end{tabular}} \\ 
\midrule
\multicolumn{6}{c}{\textbf{Results on NyayaFilter Test Data}} \\ 
\midrule
InLegalBert & \textbf{0.5870} & \textbf{0.5857} & \textbf{0.5864} & \textbf{0.5885} & \multirow{2}{*}{\begin{tabular}[c]{@{}l@{}}NyayaFacts \\ Single\end{tabular}} \\
XLNet\_Large & 0.5805 & 0.5775 & 0.5790 & 0.5818 & \\ 
\midrule
InLegalBert & 0.5886 & 0.5560 & 0.5719 & 0.5421 & \multirow{2}{*}{\begin{tabular}[c]{@{}l@{}}NyayaFacts \\ Multi\end{tabular}} \\
XLNet\_Large & \textbf{0.5977} & \textbf{0.5874} & \textbf{0.5925} & \textbf{0.5797} & \\ 
\midrule
InLegalBert & 0.5342 & 0.5180 & 0.5260 & 0.5023 & \multirow{2}{*}{\begin{tabular}[c]{@{}l@{}}NyayaScrape \\ Single\end{tabular}} \\
XLNet\_Large & \textbf{0.5577} & \textbf{0.5509} & \textbf{0.5543} & \textbf{0.5429} & \\ 
\midrule
InLegalBert & \textbf{0.5789} & \textbf{0.5409} & \textbf{0.5592} & \textbf{0.5249} & \multirow{2}{*}{\begin{tabular}[c]{@{}l@{}}NyayaScrape \\ Multi\end{tabular}} \\
XLNet\_Large & 0.5581 & 0.5364 & 0.5470 & 0.5224 & \\ 
\midrule
\multicolumn{6}{c}{\textbf{Results on NyayaSimplify Test Data}} \\ 
\midrule
InLegalBert & \textbf{0.6199} & \textbf{0.6197} & \textbf{0.6198} & 0.6167 & \multirow{2}{*}{\begin{tabular}[c]{@{}l@{}}NyayaFacts \\ Single\end{tabular}} \\
XLNet\_Large & 0.6179 & 0.6169 & 0.6174 & \textbf{0.6200} & \\ 
\midrule
InLegalBert & \textbf{0.6222} & 0.5986 & \textbf{0.6102} & 0.5839 & \multirow{2}{*}{\begin{tabular}[c]{@{}l@{}}NyayaFacts \\ Multi\end{tabular}} \\
XLNet\_Large & 0.6160 & \textbf{0.6002} & 0.6080 & \textbf{0.5878} & \\ 
\midrule
InLegalBert & 0.5760 & 0.5311 & 0.5526 & 0.5061 & \multirow{2}{*}{\begin{tabular}[c]{@{}l@{}}NyayaScrape \\ Single\end{tabular}} \\
XLNet\_Large & \textbf{0.5864} & \textbf{0.5845} & \textbf{0.5854} & \textbf{0.5789} & \\ 
\midrule
InLegalBert & 0.5659 & 0.5215 & 0.5428 & 0.4950 & \multirow{2}{*}{\begin{tabular}[c]{@{}l@{}}NyayaScrape \\ Multi\end{tabular}} \\
XLNet\_Large & \textbf{0.5978} & \textbf{0.5891} & \textbf{0.5934} & \textbf{0.5789} & \\ 
\bottomrule
\end{tabular}
}
\caption{Model performance on NyayaFilter and NyayaSimplify test datasets. For NyayaFilter, results illustrate how automatically retrieved factual data affects performance when models are trained on NyayaFacts or NyayaScrape datasets. For NyayaSimplify, results show the impact of paraphrasing complex legal texts into simpler language.}
\label{tab:nyaya_combined_results_updated}
\end{table}
\begin{table*}[t]
\centering
\resizebox{0.83\linewidth}{!}{%
\begin{tabular}{llrrrrrrc}
\toprule
& &
\multicolumn{4}{c}{\textbf{Lexical Based Evaluation}} &
\multicolumn{2}{c}{\textbf{Semantic Evaluation}}  \\ \cline{3-8}
\multirow{-2}{*}{\begin{tabular}[c]{@{}c@{}}\textbf{Training} \\ \textbf{Data}\end{tabular}} & 
\multirow{-2}{*}{\begin{tabular}[c]{@{}c@{}}\textbf{Testing} \\ \textbf{Data}\end{tabular}} & \textbf{R1} & \textbf{R2} & \textbf{RL} & \textbf{BLEU} & \textbf{BERTScore} & \textbf{BLANC} &
\multirow{-2}{*}{\begin{tabular}[c]{@{}c@{}}\textbf{Expert} \\ \textbf{Score}\end{tabular}} \\
\midrule
No Training        & NyayaFacts     & 0.28 & 0.10 & 0.14 & 0.04 & 0.53 & 0.08 &  4.2 \\
No Training        & NyayaScrape    & 0.19 & 0.08 & 0.13 & 0.04 & 0.48 & 0.09 &  4.0  \\
NyayaFacts Single  & NyayaFacts     & 0.32 & 0.11 & 0.19 & 0.04 & 0.58 & 0.10 & 7.6  \\
NyayaFacts Multi   & NyayaFacts     & \textbf{0.34} & \textbf{0.11} & \textbf{0.20} & \textbf{0.05} & \textbf{0.58} & \textbf{0.10} & \textbf{8.3}  \\
NyayaScrape Single & NyayaScrape    & 0.12 & 0.05 & 0.09 & 0.02 & 0.39 & 0.06 & 4.9  \\
NyayaScrape Multi  & NyayaScrape    & 0.17 & 0.08 & 0.13 & 0.03 & 0.45 & 0.08 & 5.6  \\
NyayaSimplify      & NyayaSimplify  & 0.28 & 0.08 & 0.18 & 0.02 & 0.56 & 0.07 & 6.7  \\ 
\bottomrule
\end{tabular}
}
\caption{Performance of FactLegalLlama on the FJPE task. The base model is LLaMa-3-8B. "No Training" indicates results from the unmodified (vanilla) model. Other rows show improvements after fine-tuning with different subsets of the \texttt{TathyaNyaya} data. }
\label{tab:factlegal_performance}
\end{table*}

In this section, we present and interpret the performance of our models across various datasets and experimental settings. We focus first on raw judgment prediction results using {NyayaFacts} and {NyayaScrape} data, then on the performance improvements or trade-offs observed in the {NyayaFilter} and {NyayaSimplify} settings. Finally, we analyze the explanation quality generated by using lexical, semantic, and expert evaluation metrics.

\subsection{Performance on NyayaFacts and NyayaScrape}
We begin by examining model performances on the \texttt{NyayaFacts} and \texttt{NyayaScrape} test sets, as reported in Table~\ref{tab:merged_results_test}. Each model
was evaluated under different training configurations, including Single and Multi.

\paragraph{Language Model-Based Baselines:}  
Across both {NyayaFacts} and {NyayaScrape} test sets, XLNet large consistently outperforms InLegalBERT. For instance, when trained on {NyayaFacts Single}, XLNet achieves a macro F1 of 0.6052 and Accuracy of 0.6061, surpassing InLegalBERT’s macro F1 of 0.5935 and Accuracy of 0.5932. This trend persists in most training and testing configurations, highlighting XLNet’s robust capability for factual judgment prediction in the given domain.

\paragraph{FactLegalLlama’s Prediction-Only Performance:}  
{FactLegalLlama}, while instruction-tuned for outcome prediction, lags behind the transformer-based baselines in raw prediction performance. For example, when trained on {NyayaFacts Single} and tested on {NyayaFacts}, it obtains a macro F1 of 0.5036 compared to XLNet\_Large’s 0.6052. A similar gap is observed across other splits. 

\paragraph{Single vs. Multi Cases:}  
Both baselines and FactLegalLlama exhibit more stable performance on the Single subsets compared to the Multi subsets. The complexity introduced by multiple petitions with varying outcomes in the Multi cases reduces overall accuracy and F1 scores, emphasizing the challenge of fact-based judgment prediction in more intricate legal scenarios.

\subsection{Impact of Fact Retrieval (NyayaFilter) and Text Simplification (NyayaSimplify)}
Table~\ref{tab:nyaya_combined_results_updated} reports model performances on the \texttt{NyayaFilter} and \texttt{NyayaSimplify} test datasets. These results highlight how the preprocessing choices affect model accuracy on automatic fact retrieval and paraphrasing complex legal texts.

\paragraph{NyayaFilter Results:}  
When comparing \texttt{NyayaFilter} results to the original \texttt{NyayaFacts} and \texttt{NyayaScrape} sets, we see that while performance can fluctuate, some models benefit from training on data where fact and non-fact segments are clearly distinguished. For example, on the \texttt{NyayaFilter} test set derived from \texttt{NyayaFacts Single}, InLegalBERT attains a macro F1 of 0.5864, maintaining competitive performance. 
These findings suggest that automatically retrieved factual subsets can be used without severely degrading model performance.

\paragraph{NyayaSimplify Results:}  
Paraphrasing complex legal language into simpler text (the \texttt{NyayaSimplify} scenario) generally helps models retain or slightly improve performance. For instance, with \texttt{NyayaFacts Single}, InLegalBERT reaches a macro F1 of 0.6198 and XLNet\_Large hits an Accuracy of 0.6200 on the simplified data, both representing small yet noteworthy improvements compared to their performance on the original complex texts. This trend indicates that reducing linguistic complexity can aid models in understanding and classifying factual statements better.
\begin{table*}[t]
\centering
\resizebox{0.9\linewidth}{!}{%
\begin{tabular}{lllccccc}
\toprule
\textbf{Base Model} & \textbf{Training Data} & \textbf{Testing Data} & \textbf{Fleiss' $\kappa$} & \textbf{Cohen's $\kappa$} & \textbf{ICC} & \textbf{Kripp. $\alpha$} & \textbf{Pearson Corr.} \\
\midrule
Meta-LLaMA-3-8B & No Training & NyayaFacts       & 0.62 & 0.59 & 0.64 & 0.61 & 0.65 \\
Meta-LLaMA-3-8B & No Training & NyayaScrape      & 0.54 & 0.50 & 0.55 & 0.53 & 0.56 \\
Meta-LLaMA-3-8B & NyayaFacts Single & NyayaFacts   & 0.70 & 0.68 & 0.71 & 0.69 & 0.72 \\
Meta-LLaMA-3-8B & NyayaFacts Multi & NyayaFacts    & \textbf{0.81} & \textbf{0.79} & \textbf{0.83} & \textbf{0.80} & \textbf{0.84} \\
Meta-LLaMA-3-8B & NyayaScrape Single & NyayaScrape & 0.58 & 0.54 & 0.60 & 0.57 & 0.59 \\
Meta-LLaMA-3-8B & NyayaScrape Multi & NyayaScrape  & 0.63 & 0.60 & 0.66 & 0.62 & 0.64 \\
Meta-LLaMA-3-8B & NyayaSimplify & NyayaSimplify   & 0.69 & 0.66 & 0.70 & 0.68 & 0.70 \\
\bottomrule
\end{tabular}
}
\caption{Inter-Annotator Agreement (IAA) metrics for expert evaluation across different training and testing setups. Higher scores indicate stronger agreement and reliability of expert-based assessments.}
\label{tab:iaa_metrics}
\end{table*}

\subsection{Quality of Explanations}
Table~\ref{tab:factlegal_performance} presents the evaluation of {FactLegalLlama} on the explanation generation task, measured through lexical, semantic, and expert evaluation metrics. We compare a "No Training" scenario with fine-tuned versions of \texttt{TathyaNyaya} data.

\paragraph{Fine-tuning Benefits:}  
\texttt{FactLegalLlama} on factual data substantially improves its explanation quality. For \texttt{NyayaFacts}, training on the Multi subset yields the strongest results, outperforming both the "No Training" scenario and the Single subset training. This suggests that exposure to more complex, multi-petition cases helps the model generate richer, more contextually sensitive explanations.

\paragraph{Domain-Specific Fine-tuning:}  
The contrast between "No Training" and the various training configurations highlights the necessity of domain-specific adaptation. Without fine-tuning, the model’s explanations remain weak and less aligned with factual inputs, as indicated by lower Rouge and BLEU scores. After training with \texttt{NyayaFacts Multi}, the model better captures the underlying legal rationale, producing explanations that align more closely with reference annotations.

\paragraph{Expert Evaluation:}
As shown in Table~\ref{tab:factlegal_performance}, we complemented our automatic lexical and semantic evaluation with a human-centric expert assessment using a 1–10 Likert scale. Legal experts rated the generated explanations on clarity, legal relevance, and factual consistency. Results demonstrate that \texttt{FactLegalLlama} fine-tuned on \texttt{NyayaFacts Multi}, achieved the highest score (average 8.3), reflecting strong alignment with human-annotated rationales. Fine-tuning on simplified data (\texttt{NyayaSimplify}) and single-case subsets also yielded substantial improvements (6.7 and 7.6, respectively), while explanations from the zero-shot model (No Training) were rated notably lower (around 4.0). These findings highlight the importance of domain-specific supervision and exposure to complex multi-petition cases for producing high-quality, interpretable explanations.

To ensure reliability of expert ratings, we further computed IAA results, summarized in Table~\ref{tab:iaa_metrics}, which indicate moderate to high agreement across evaluators, with the strongest consistency observed for \texttt{NyayaFacts Multi}. This demonstrates that expert judgments were stable and reproducible, reducing subjectivity and strengthening the credibility of our evaluation framework. Together, the expert scores and IAA results confirm that improvements in explanation quality are not only measurable but also consistently recognized by multiple evaluators.

\section{Conclusions and Future Work}
\label{sec:conclusions_future_work}

We introduced \texttt{TathyaNyaya}, a fact-focused dataset for judgment prediction and explanation within the Indian legal domain, and \texttt{FactLegalLlama}, an instruction-tuned model delivering fact-grounded rationales. By emphasizing factual content rather than full judgments, \texttt{TathyaNyaya} aligns more closely with actual legal decision-making scenarios, while \texttt{FactLegalLlama} highlights the value of coupling predictive accuracy with transparent explanations. Preprocessing steps such as fact filtering and paraphrasing further enhance model clarity and performance, and domain-specific fine-tuning proves essential for capturing legal subtleties. Future work may extend these findings to other jurisdictions.

\section*{Acknowledgements}

We would like to express our sincere gratitude to the anonymous reviewers for their constructive feedback and valuable suggestions, which greatly improved the quality of this paper. We also thank the student research assistants and legal experts from various law colleges for their dedicated efforts in annotation and expert evaluation. Their contributions were instrumental in ensuring the quality and reliability of the dataset and evaluation process.
We gratefully acknowledge the support of BharatGen, India, for providing access to computational resources and hardware infrastructure used in this research. Their assistance played a vital role in model training and large-scale experimentation. The majority of this work was conducted while the first author was affiliated with the Indian Institute of Technology Kanpur. The author is currently affiliated with the University of Birmingham Dubai.

\section*{Limitations}
This study faced several limitations that influenced both the scope and outcomes of our research. A key constraint was the reliance on a 4-bit quantized model due to resource limitations, which restricted our ability to experiment with larger parametric models, such as 70B or 40B parameter LLMs. Additionally, the high computational costs and token limitations associated with cloud-based services further hindered our capacity to perform extensive inference and fine-tuning. This restricted exploration may have limited the depth of insights and performance metrics achievable with FactLegalLlama.


The dataset used in this study comprises only English-language judgments, which limits its applicability in multilingual contexts, especially in jurisdictions where regional languages dominate legal proceedings. This exclusion highlights the need for more inclusive datasets that reflect the linguistic diversity of legal documents in India and beyond.

These limitations underscore the challenges of applying LLMs to specialized legal tasks such as judgment prediction and explanation. They also point to areas requiring further research, including resource optimization, multilingual dataset development, and enhancing the factual consistency and reasoning capabilities of AI models.

\section*{Ethics Statement}
This research was conducted with a strong commitment to ethical considerations, particularly given the sensitive nature of legal data and the implications of deploying AI in legal contexts. The \texttt{TathyaNyaya} dataset, central to this study, was compiled from publicly accessible sources, such as Indian legal search engines, ensuring adherence to data privacy and usage regulations. To further safeguard privacy, we removed identifiable meta-information, including judge names, case titles, and case IDs, from the dataset.

The computational resources used for model training and evaluation were obtained through ethical and legitimate means. These resources were either institutional or subscribed services, ensuring compliance with licensing agreements and financial support for these platforms. By adhering to these practices, we ensured that our research activities aligned with sustainable and lawful resource usage.

Transparency and reproducibility were foundational principles of this study. The \texttt{TathyaNyaya} dataset and the code for FactLegalLlama will be made publicly available, enabling researchers to replicate and extend our findings. This open-access approach is intended to foster collaboration within the research community and drive further advancements in AI-assisted legal decision-making.

We recognize the potential societal impact of AI applications in the legal domain, particularly regarding fairness, accountability, and the risk of misuse. Our models are explicitly designed to assist legal professionals rather than replace human judgment, emphasizing the necessity of human oversight in AI-assisted decision-making processes. As we continue this line of research, we remain vigilant in addressing ethical challenges and aligning our efforts with principles of fairness, transparency, and societal benefit.

\newpage
\bibliography{ custom}

\newpage
\appendix
\section{Experimental Setup and Hyper-parameters}
\label{sec:Experimental-setup}
In this section, we detail the experimental configurations, training procedures, and hyper-parameters employed to develop and evaluate our models. We first describe the training of transformer-based baseline models for fact-based judgment prediction, then outline the instruction-tuning process used to adapt \texttt{FactLegalLlama} for both prediction-only and prediction-with-explanation tasks.

\subsection{Transformers Training Hyper-parameters}
To establish competitive baselines, we fine-tuned transformer models such as InLegalBERT and XLNet\_Large on the \texttt{NyayaFacts} dataset. Each model was trained with a batch size of 16 using the AdamW optimizer \cite{kingma2014adam} and a learning rate of 2e-6. We ran the training for three epochs, adopting default hyper-parameter settings from the HuggingFace Transformers library. Experiments were carried out on an NVIDIA A100 40GB GPU, ensuring adequate computational resources for handling extensive legal text. This training protocol allowed the models to capture the nuances of fact-based segments and reliably predict judicial outcomes.

\subsection{FactLegalLlama Instruction Fine-Tuning} 
To develop \texttt{FactLegalLlama}, we began with the meta-llama/Meta-Llama-3-8B base model. We applied 4-bit quantization to optimize memory usage and introduced Low-Rank Adaptation (LoRA) with a rank of 16 for parameter-efficient fine-tuning. The maximum input sequence length was set to 2,500 tokens, accommodating the substantial factual inputs characteristic of legal documents.

We employed the paged AdamW optimizer in 32-bit precision with a learning rate of 1e-4 and implemented a cosine decay learning rate scheduler for smoother convergence. Mixed-precision training (fp16) and a gradient accumulation of 4 steps were used to further manage GPU memory. We utilized a per-device batch size of 4 and trained the model for three epochs, a process that required approximately 38 hours on an NVIDIA A100 40GB GPU. Under these conditions, the model achieved a training loss of 1.5060 and a validation loss of 1.6745, indicating effective adaptation to the underlying factual patterns in the data.

\subsection{Training Objectives} The instruction-based fine-tuning of \texttt{FactLegalLlama} targeted two primary objectives: fact-driven judgment prediction and fact-driven prediction with explanation. By employing a carefully designed set of instructions and incorporating LoRA-based parameter updates, the model learned to generate outcomes and accompanying rationales rooted in the factual segments. This combination of parameter-efficient fine-tuning and instruction-oriented training yielded a model well-suited for practical applications in legal NLP, balancing computational feasibility with interpretability and domain relevance.

\subsection{Training Procedure for Hierarchical BiLSTM-CRF Classifier}
\label{sec:training_procedure}

The Hierarchical BiLSTM-CRF classifier is designed to classify sentences in legal documents into factual and non-factual categories by leveraging the hierarchical structure of the data. The model architecture comprises a word-level BiLSTM coupled with a CRF layer and a sentence-level BiLSTM. The word-level BiLSTM encodes contextual dependencies within sentences, while the CRF ensures coherence in predicted tag sequences. The sentence-level BiLSTM aggregates these representations to capture inter-sentence dependencies, enabling the model to account for both local and global patterns in the data.

Training is conducted using the AdamW optimizer with a learning rate of 2e-6, a batch size of 16, and for five epochs. A CRF-based loss function is used to optimize sequence-level tagging accuracy. During training, metrics such as precision, recall, F1-score, and loss are evaluated on a validation set after each epoch to monitor performance and ensure generalization. The model configuration includes a word embedding size of 100 and a sentence embedding size of 200, with training conducted on an NVIDIA A100 40GB GPU.

To enhance generalization, K-fold cross-validation is employed, where the dataset is split into multiple folds, and the model is trained and validated on different subsets. The average performance across folds provides a robust measure of the model's capability. Checkpoints are saved periodically during training, enabling the model to be restored for inference or further fine-tuning.

\begin{table*}[ht]
    \centering
    \begin{tabular}{|p{0.8\textwidth}|}
    \hline
    {\bf Template 1 (prediction only)}\\
    \hline
    {\bf prompt} = f``````
    \#\#\# {\bf Instructions}: Given the facts of the case,just predict the outcome as '1' for acceptance or '0' for rejection. \\
    
    \#\#\# \textbf{Input}: $<$\{case\_facts\}$>$\\
    
    \#\#\# Response:
    ''''''\\
    \hline
    {\bf Template 2 (prediction with explanation)}\\
    \hline
    {\bf prompt} = f``````
    \#\#\# {\bf Instructions}: Given the facts of the case,first predict the outcome as '1' for acceptance or '0' for rejection. Then, provide key sentences from the facts or clear reasoning that support your decision. \\
    
    \#\#\# \textbf{Input}: $<$\{case\_facts\}$>$\\
    
    \#\#\# Response:
    ''''''\\
    \hline
    \end{tabular}
    \caption{Prompts for Factual Judgment Prediction and Explanation used for instruction fine-tuned models. Instructions were selected based on the templates provided in Table \ref{Instruction-sets}.}
    \label{tab:factual_judgment_prediction_prompts}
\end{table*}
\begin{table*}[ht]
    \centering
    \begin{tabular}{|p{0.8\textwidth}|}
    \hline
    {\bf Template 1 (Paraphrasing facts)}\\
    \hline
    {\bf prompt} = f``````
    \#\#\# {\bf Instructions}:You are an Indian legal expert with extensive knowledge of legal terms, statutes, and laws. Your task is to explain a legal case to your clients in simple and understandable language.
    Avoid legal jargon and focus on conveying the meaning of the case in everyday language, making it clear and easy for someone without legal knowledge to understand.
    While simplifying, ensure that the key points of the case, including the facts, legal claims, and decisions, are clearly communicated without losing any critical information.
    You should Preserve the key legal terms and references,Clarify complex legal processes,Avoid excessive legal jargon,Be concise but complete,Explain court actions clearly, Provide Only Paraphrased Outcome\\
    
    \#\#\# \textbf{Input}: Paraphrase the following text:$<$\{case\_facts\}$>$\\
    
    \#\#\# Response:
    ''''''\\
    \hline
    
    \end{tabular}
    \caption{Prompt for paraphrasing facts to change legal jargons to interpretable terms.}
    \label{tab:nyayasimplify_prompt}
\end{table*}
\begin{figure*}[ht]
    \centering
    \includegraphics[width=0.8\linewidth]{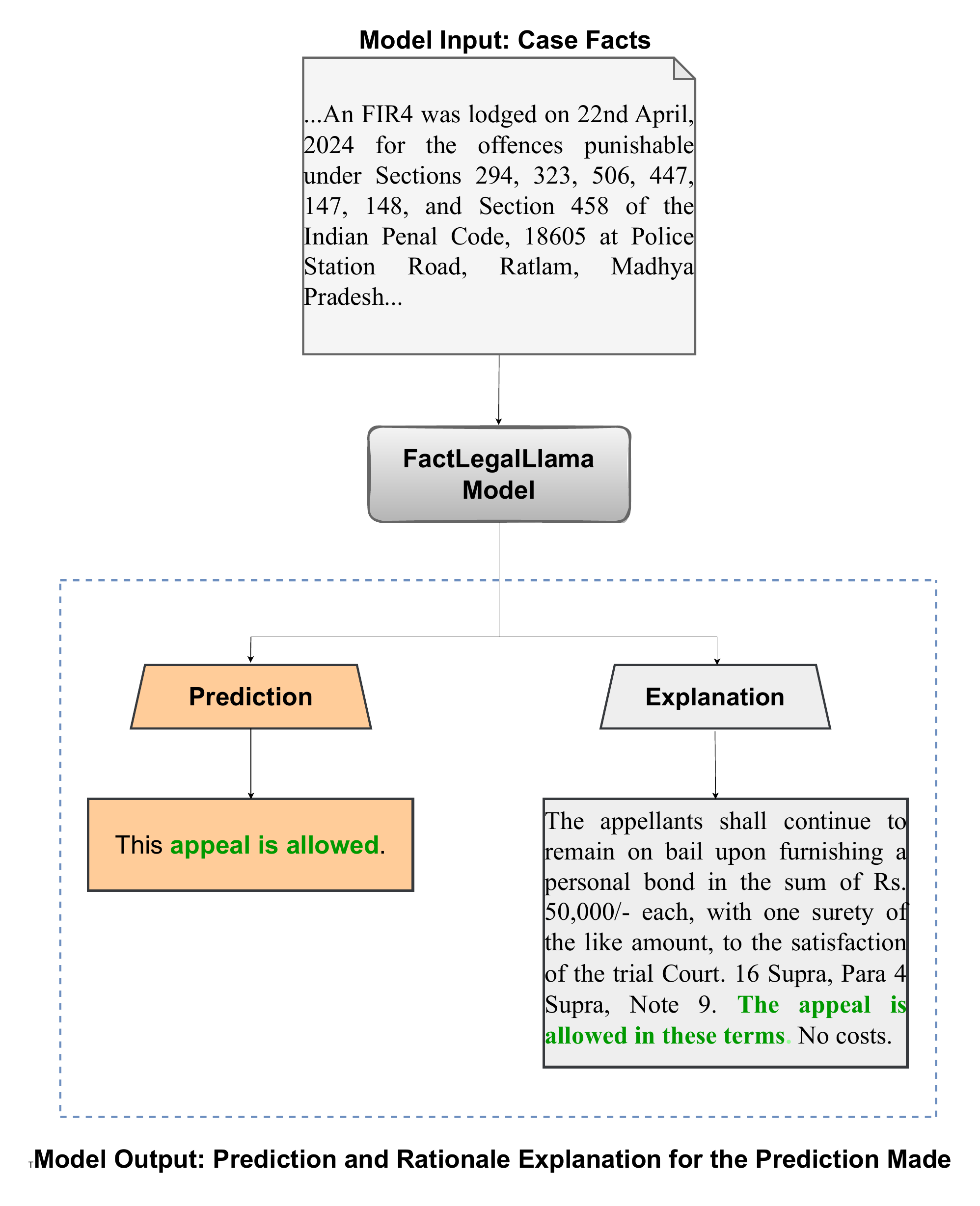}
    \caption{Illustration of the Fact-based Judgment Prediction and Explanation (FJPE) pipeline using the \texttt{FactLegalLlama} model.}
    \label{fig:fjpe_task_framework}
\end{figure*}

\begin{figure*}[ht]
    \centering
    \includegraphics[width=\linewidth]{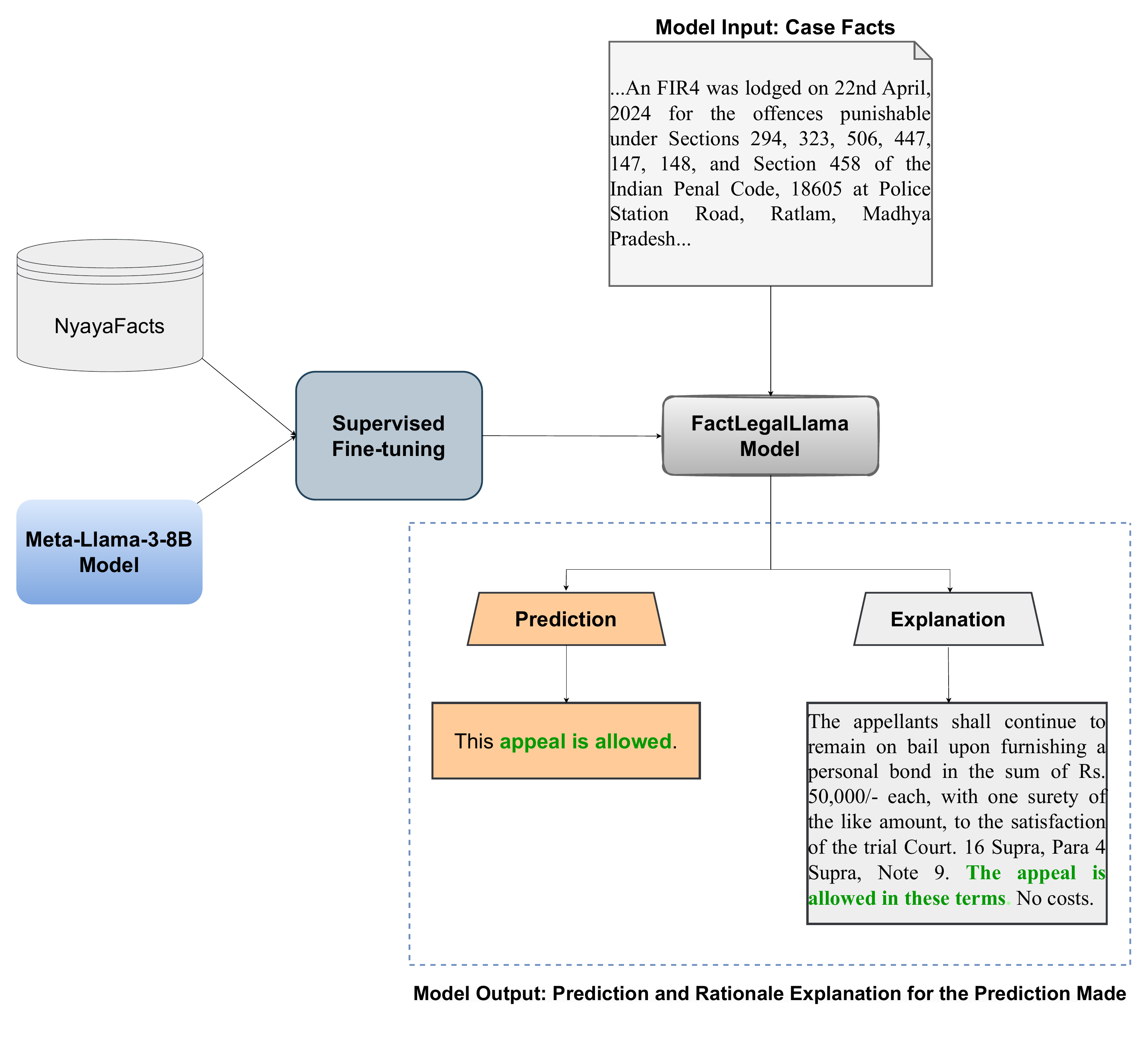}
    \caption{Training dynamics of \texttt{FactLegalLlama} for the combined judgment prediction and explanation task. The model learns to produce both the outcome and its underlying rationale directly from factual inputs, guided by instruction-based fine-tuning.}
    \label{fig:factlegalllama_training}
\end{figure*}

\begin{figure*}[ht]
    \centering
    \includegraphics[width=\linewidth]{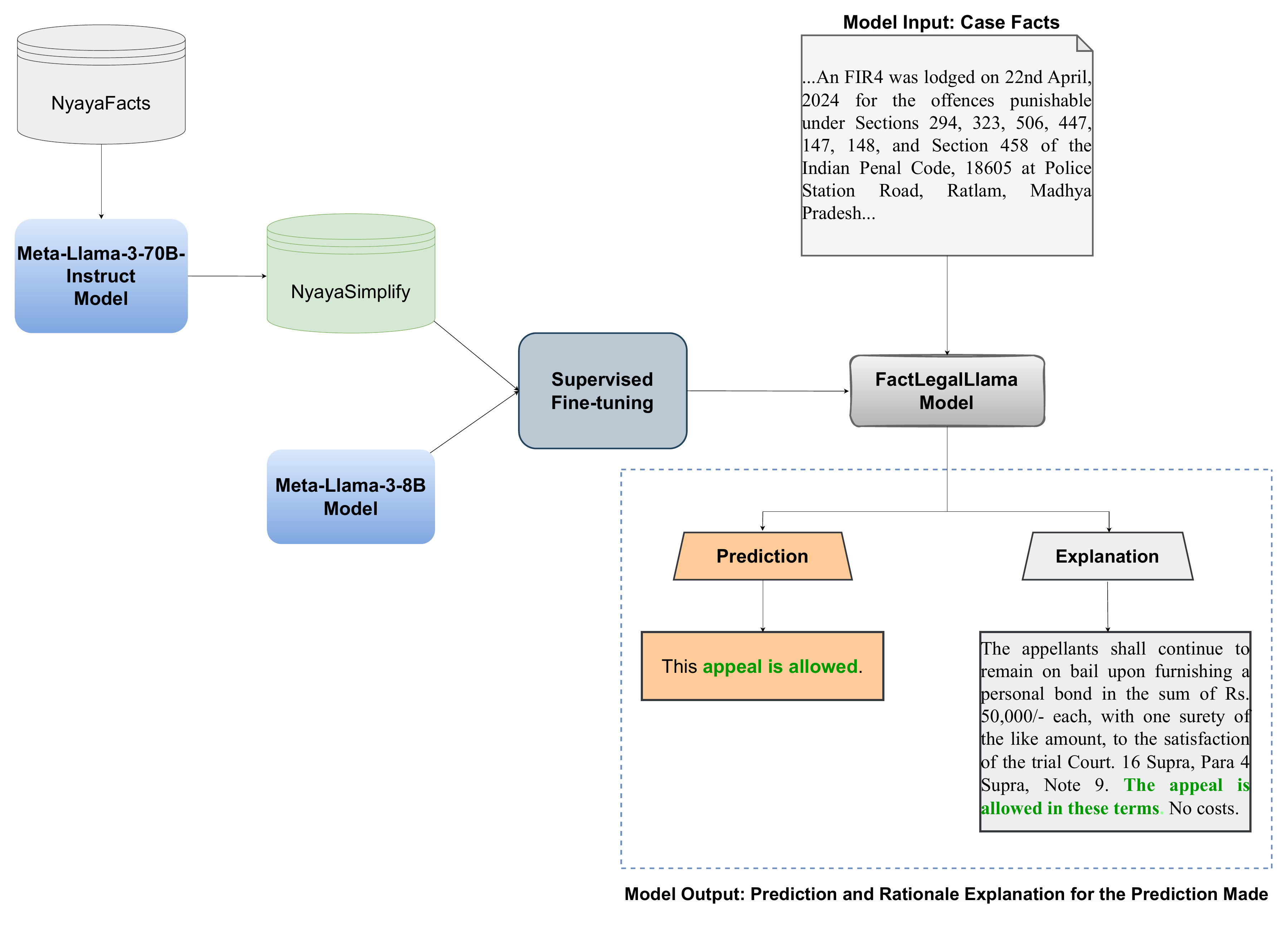}
    \caption{Overview of the simplification and fine-tuning process. First, complex legal facts are paraphrased into simpler language using LLaMA-3-70B, creating the \texttt{NyayaSimplify} dataset, followed by supervised fine-tuning (SFT) using LLaMa-3-7B for the FJPE task.}
    \label{fig:paraphrased_sft}
\end{figure*}

\begin{figure*}[ht]
    \centering
    \includegraphics[width=\linewidth]{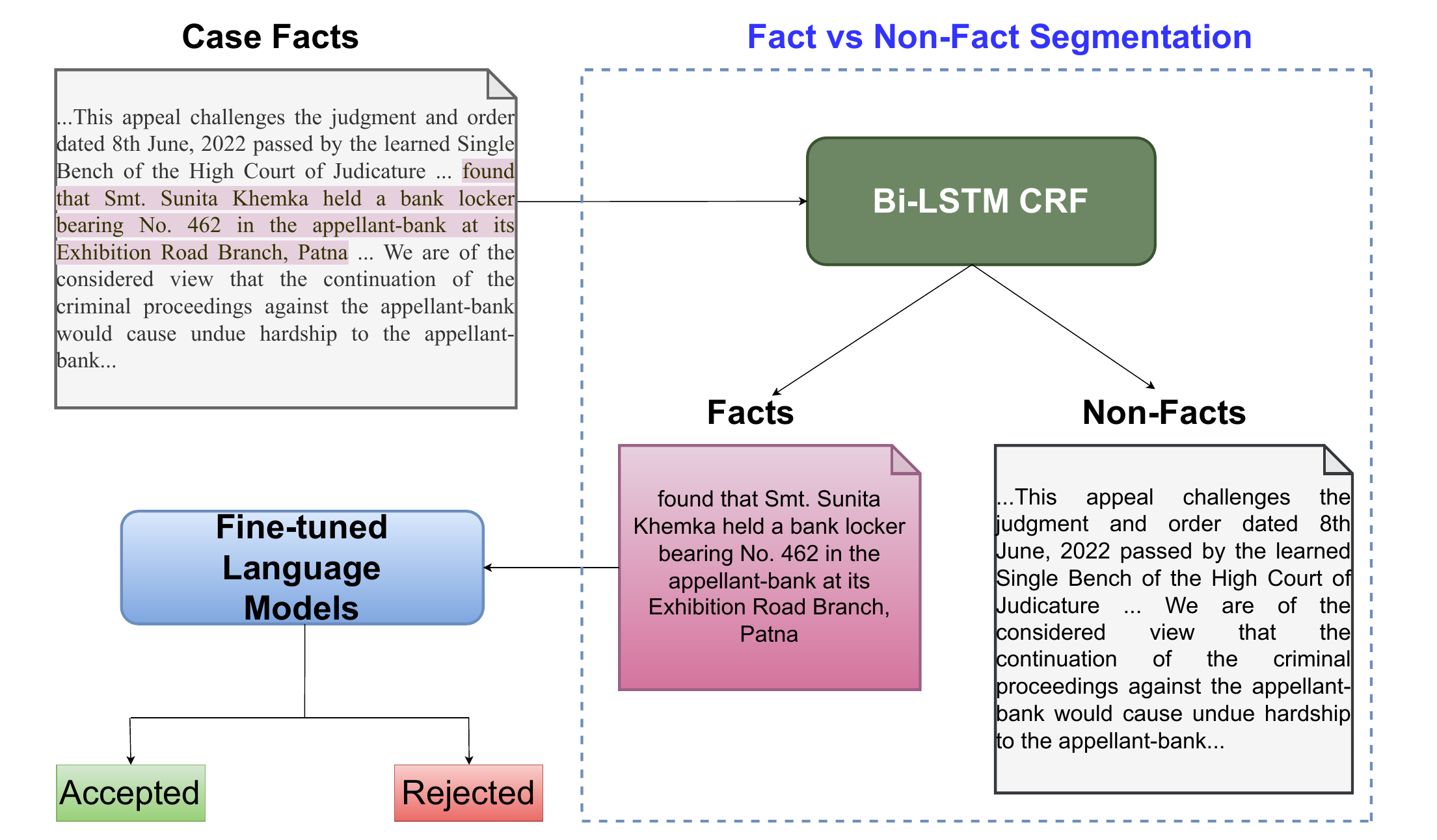}
    \caption{The Fact vs. Non-Fact segmentation framework employing a BiLSTM-CRF model. This segmentation step separates factual statements from non-factual content in legal judgments, creating the \texttt{NyayaFilter} dataset. The refined dataset is subsequently used for downstream judgment prediction and explanation tasks.}

    \label{fig:fact_vs_non_fact_segmentation}
\end{figure*}

\begin{table*}[ht]
\centering
\resizebox{0.85\textwidth}{!}{%
\tiny
\begin{tabular}{|cl|}
\hline
\multicolumn{2}{|c|}{\textbf{\textcolor{blue}{Instruction sets for Predicting the Decision}}} \\ \hline

\multicolumn{1}{|c|}{1} &
  \begin{tabular}[c]{@{}l@{}}Analyze the facts presented in the case and predict whether the outcome will be favorable (1) or \\unfavorable (0).\end{tabular} \\ \hline
  
\multicolumn{1}{|c|}{2} &
  \begin{tabular}[c]{@{}l@{}}Based on the facts provided, determine the likely outcome: favorable (1) or unfavorable (0) for the \\appellant/petitioner\end{tabular} \\ \hline
  
\multicolumn{1}{|c|}{3} &
  \begin{tabular}[c]{@{}l@{}}Review the facts of the case and predict the decision: will the court rule in favor (1) or against (0) the \\appellant/petitioner?\end{tabular} \\ \hline
  
\multicolumn{1}{|c|}{4} &
  \begin{tabular}[c]{@{}l@{}}Considering the facts and evidence in the case, predict the verdict: is it more likely to be in favor (1) or \\against (0) the appellant?\end{tabular} \\ \hline
  
\multicolumn{1}{|c|}{5} &
  \begin{tabular}[c]{@{}l@{}}Examine the facts of the case and forecast whether the appeal/petition is likely to be upheld (1) or \\dismissed (0).\end{tabular} \\ \hline
  
\multicolumn{1}{|c|}{6} &
  \begin{tabular}[c]{@{}l@{}}Assess the facts of the case and provide a prediction: is the court likely to rule in favor of (1) or \\against (0) the appellant/petitioner?\end{tabular} \\ \hline
  
\multicolumn{1}{|c|}{7} &
  \begin{tabular}[c]{@{}l@{}}Interpret the facts of the case and speculate on the court's decision: will the appeal be accepted (1) or \\rejected (0) based on the provided information?\end{tabular} \\ \hline
  
\multicolumn{1}{|c|}{8} &
  \begin{tabular}[c]{@{}l@{}}Given the specifics of the case facts, anticipate the court's ruling: will it favor (1) or oppose (0) the \\appellant’s request?\end{tabular} \\ \hline
  
\multicolumn{1}{|c|}{9} &
  \begin{tabular}[c]{@{}l@{}}Scrutinize the facts and arguments presented in the case to predict the court's decision: will the appeal \\be granted (1) or denied (0)?\end{tabular} \\ \hline
  
\multicolumn{1}{|c|}{10} &
  \begin{tabular}[c]{@{}l@{}}Analyze the facts presented and estimate the likelihood of the court accepting (1) or rejecting (0) the \\petition.\end{tabular} \\ \hline
  
\multicolumn{1}{|c|}{11} &
  \begin{tabular}[c]{@{}l@{}}From the facts provided in the case, infer whether the court's decision will be favorable (1) or \\unfavorable (0) for the appellant.\end{tabular} \\ \hline
  
\multicolumn{1}{|c|}{12} &
  \begin{tabular}[c]{@{}l@{}}Evaluate the facts and evidence in the case and predict the verdict: is an acceptance (1) or rejection \\(0) of the appeal more probable?\end{tabular} \\ \hline
  
\multicolumn{1}{|c|}{13} &
  \begin{tabular}[c]{@{}l@{}}Delve into the case facts and predict the outcome: is the judgment expected to be in support (1) or in \\denial (0) of the appeal?\end{tabular} \\ \hline
  
\multicolumn{1}{|c|}{14} &
  \begin{tabular}[c]{@{}l@{}}Using the case facts, forecast whether the court is likely to side with (1) or against (0) the appellant\\/petitioner.\end{tabular} \\ \hline
  
\multicolumn{1}{|c|}{15} &
  \begin{tabular}[c]{@{}l@{}}Examine the case facts and anticipate the court's decision: will it result in an approval (1) or \\disapproval (0) of the appeal?\end{tabular} \\ \hline
  
\multicolumn{1}{|c|}{16} &
  \begin{tabular}[c]{@{}l@{}}Based on the facts and evidence in the case, predict the court's stance: favorable (1) or unfavorable \\(0) to the appellant.\end{tabular} \\ \hline

\multicolumn{2}{|c|}{\textbf{\textcolor{blue}{Instruction sets for Integrated Approach for Prediction and Explanation}}} \\ \hline
\multicolumn{1}{|c|}{1} &
  \begin{tabular}[c]{@{}l@{}}First, predict whether the appeal in case proceeding will be accepted (1) or not (0), and then explain the \\decision by identifying crucial sentences from the document.\end{tabular} \\ \hline
\multicolumn{1}{|c|}{2} &
  \begin{tabular}[c]{@{}l@{}}Determine the likely decision of the case facts (acceptance (1) or rejection (0)) and follow up with \\an explanation highlighting key sentences that support this prediction.\end{tabular} \\ \hline
\multicolumn{1}{|c|}{3} &
  \begin{tabular}[c]{@{}l@{}}Predict the outcome of the case based on the facts provided (acceptance (1) or rejection (0)) and \\explain your reasoning by extracting key sentences that justify the decision.\end{tabular} \\ \hline
\multicolumn{1}{|c|}{4} &
  \begin{tabular}[c]{@{}l@{}}Evaluate the case facts to forecast the court's decision (1 for yes, 0 for no), and elucidate the \\reasoning behind this prediction with important textual evidence from the case.\end{tabular} \\ \hline
\multicolumn{1}{|c|}{5} &
  \begin{tabular}[c]{@{}l@{}}Ascertain if the court will uphold (1) or dismiss (0) the appeal based on the case facts, and then \\clarify this prediction by discussing the critical sentences that support the decision.\end{tabular} \\ \hline
\multicolumn{1}{|c|}{6} &
  \begin{tabular}[c]{@{}l@{}}Judge the probable resolution of the case based on the facts (approval (1) or disapproval (0)), and \\elaborate on this forecast by extracting and interpreting significant sentences from the case facts.\end{tabular} \\ \hline
\multicolumn{1}{|c|}{7} &
  \begin{tabular}[c]{@{}l@{}}Forecast the likely verdict of the case (granting (1) or denying (0) the appeal) based on the facts, \\and rationalize your prediction by pinpointing and explaining pivotal sentences in the case document.\end{tabular} \\ \hline
\multicolumn{1}{|c|}{8} &
  \begin{tabular}[c]{@{}l@{}}Assess the case to predict the court's ruling (favorably (1) or unfavorably (0)) based on the facts, \\and expound on this prediction by highlighting and analyzing key textual elements from the case facts.\end{tabular} \\ \hline
\multicolumn{1}{|c|}{9} &
  \begin{tabular}[c]{@{}l@{}}Assess the case to predict the court's ruling (favorably (1) or unfavorably (0)) based on the facts, \\and expound on this prediction by highlighting and analyzing key textual elements from the case facts.\end{tabular} \\ \hline
\multicolumn{1}{|c|}{10} &
  \begin{tabular}[c]{@{}l@{}}Conjecture the end result of the case (acceptance (1) or non-acceptance (0) of the appeal) based \\on the facts, followed by a detailed explanation using crucial sentences from the case facts.\end{tabular} \\ \hline
\multicolumn{1}{|c|}{11} &
  \begin{tabular}[c]{@{}l@{}}Predict whether the case will result in an affirmative (1) or negative (0) decision for the appeal based \\on the facts, and then provide a thorough explanation using key sentences to support your prediction.\end{tabular} \\ \hline
\multicolumn{1}{|c|}{12} &
  \begin{tabular}[c]{@{}l@{}}Estimate the outcome of the case (positive (1) or negative (0) for the appellant) based on the facts, and \\then provide a reasoned explanation by examining important sentences within the case documentation.\end{tabular} \\ \hline
\multicolumn{1}{|c|}{13} &
  \begin{tabular}[c]{@{}l@{}}Project the court's decision (favor (1) or against (0) the appeal) based on the case facts, and \\subsequently provide an in-depth explanation by analyzing relevant sentences from the document.\end{tabular} \\ \hline
\multicolumn{1}{|c|}{14} &
  \begin{tabular}[c]{@{}l@{}}Make a prediction on the court's ruling (acceptance (1) or rejection (0) of the petition) based on the \\case facts, and then dissect the case to provide a detailed explanation using key textual passages.\end{tabular} \\ \hline
\multicolumn{1}{|c|}{15} &
  \begin{tabular}[c]{@{}l@{}}Speculate on the likely judgment (yes (1) or no (0) to the appeal) based on the case facts, and then \\delve into the case to elucidate your prediction, focusing on critical sentences.\end{tabular} \\ \hline
\multicolumn{1}{|c|}{16} &
  \begin{tabular}[c]{@{}l@{}}Hypothesize the court's verdict (affirmation (1) or negation (0) of the appeal) based on the case facts, \\and then clarify this hypothesis by interpreting significant sentences from the case.\end{tabular} \\ \hline

\end{tabular}%
}
\caption{Instruction sets for Prediction and Explanation using factual data from case proceedings.}
\label{Instruction-sets}
\end{table*}

\end{document}